\documentclass[conference]{IEEEtran}
\IEEEoverridecommandlockouts
\usepackage{cite}
\usepackage{amsmath,amssymb,amsfonts}
\usepackage{algorithm}  
\usepackage{algpseudocode}
\usepackage{multirow}

\usepackage{amsfonts} 

\usepackage{hyperref}
\usepackage{subcaption}
\usepackage{array} 
\usepackage{makecell} 
\usepackage[flushleft]{threeparttable}
\usepackage{tabularx} 
\usepackage{graphicx}
\usepackage{epstopdf}

\usepackage{textcomp}
\usepackage{xcolor}
\def\BibTeX{{\rm B\kern-.05em{\sc i\kern-.025em b}\kern-.08em
    T\kern-.1667em\lower.7ex\hbox{E}\kern-.125emX}}
\def\BibTeX{{\rm B\kern-.05em{\sc i\kern-.025em b}\kern-.08em
    T\kern-.1667em\lower.7ex\hbox{E}\kern-.125emX}}
    \title{ Efficient and Safe Planner for Automated Driving on Ramps Considering Unsatisfication\\
 \thanks{$^{\dagger}$ Common contribution}
 \thanks{$^{\S}$Corresponding author}%
}

\author{
\IEEEauthorblockN{1\textsuperscript{st} Qinghao Li$^{\dagger}$}
\IEEEauthorblockA{\textit{Department of Computer Science,} \\
\textit{University of Liverpool,}\\
Liverpool L69 3GJ, United Kingdom \\
psqli35@liverpool.ac.uk}
\and
\IEEEauthorblockN{2\textsuperscript{nd} Zhen Tian$^{\dagger}$}
\IEEEauthorblockA{\textit{James Watt School of Engineering,} \\
\textit{University of Glasgow,}\\
Glasgow G12 8QQ, United Kingdom \\
2620920z@student.gla.ac.uk}
\and
\IEEEauthorblockN{3\textsuperscript{th} Xiaodan Wang}
\IEEEauthorblockA{\textit{School of Engineering,} \\
\textit{Cardiff University,}\\
Cardiff CF24 3AA, United Kingdom \\
WangX223@cardiff.ac.uk}
\and
\IEEEauthorblockN{4\textsuperscript{th} Jinming Yang}
\IEEEauthorblockA{\textit{School of Computing Science,} \\
\textit{University of Glasgow,}\\
Glasgow G12 8QQ, United Kingdom \\
j.yang.8@research.gla.ac.uk}
\and
\IEEEauthorblockN{5\textsuperscript{th} Zhihao Lin$^{\S}$}
\IEEEauthorblockA{\textit{James Watt School of Engineering,} \\
\textit{University of Glasgow,}\\
Glasgow G12 8QQ, United Kingdom \\
2800400L@student.gla.ac.uk}
}
\begin{document}
\maketitle
\begin{abstract}
Automated driving on ramps presents significant challenges due to the need to balance both safety and efficiency during lane changes. This paper proposes an integrated planner for automated vehicles (AVs) on ramps, utilizing an unsatisfactory level metric for efficiency and arrow-cluster-based sampling for safety. The planner identifies optimal times for the AV to change lanes, taking into account the vehicle's velocity as a key factor in efficiency. Additionally, the integrated planner employs arrow-cluster-based sampling to evaluate collision risks and select an optimal lane-changing curve. Extensive simulations were conducted in a ramp scenario to verify the planner's efficient and safe performance. The results demonstrate that the proposed planner can effectively select an appropriate lane-changing time point and a safe lane-changing curve for AVs, without incurring any collisions during the maneuver. 
\end{abstract}


\section{Introduction}
Recently, autonomous vehicles has been widely used for different scenarios, such as in-room navigation~\cite{lin2024enhanced}, racing~\cite{tian2024efficient}, highway driving~\cite{lenz2016tactical}, and roundabout driving~\cite{lin2024conflicts}. However, autonomous driving on ramps involves interactions with human-driven vehicles (HDVs), which often requires safe decision-making capabilities of surround the autonomous vehicle (AV)~\cite{tian2025evaluating}. Proper interaction with these HDVs is crucial for safe lane-changing~\cite{alyamani2023driver}. Therefore, AVs need to evaluate the collision risks associated with nearby HDVs during lane-changing maneuvers.

During driving on ramps, a variety of HDVs pose significant disturbances to AVs. Unlike AVs, human drivers have various demands while driving, such as safety, efficiency, and comfort~\cite{hang2020integrated}. To interact effectively with surrounding HDVs, it is essential for AVs to respond to different driving intentions. A review of existing methods for intention-aware driving for AVs is summarized in~\cite{martinez2017driving}. However, many previous studies have focused solely on friendly driving itself, ignoring the specific features of the scenario. 

In recent years, game theory has been widely used in the decision-making of automated driving~\cite{9831031}. However, game theory focuses on finding equilibrium points where no player has an incentive to deviate. In dynamic traffic situations, reaching and maintaining equilibrium may not be feasible or practical. Furthermore, game theory ignores abnormal driving behaviors because all agents are assumed to be rational and act optimally to maximize their own utility. Sometimes the vehicles will drive slowly using game theory, causing long remaining on the road. To reduce the remaining time in ramp lane and improve the driving efficiency, inspired by~\cite{6570532}, a unsatisfactory level-based decision making is applied. For the selection safe lane-changing curve, an arrow-cluster-based method is used to maintain the safety during the lane-changing. The main contributions of this paper are summarized as follows:
\begin{itemize}
    \item Safe and efficient lane-changing is achieved by applying the proposed double-side planner. The proposed planner accounts for short-term lane-changing and lane-changing curve evaluation with efficiency, safety, and comfort, thereby promoting traffic flow in the ramp lane and avoiding collisions with HDVs.
    \item This paper includes simulations that account for HDVs surrounding the AV in the ramp lane, closely mirroring actual traffic conditions. Results indicate that our method achieves efficient selection of lane-changing points within a short time period, and an efficient, safe, and comfortable lane-changing process.
\end{itemize}
\begin{figure*}[t]
    \centering
    \includegraphics[width=0.8\linewidth]{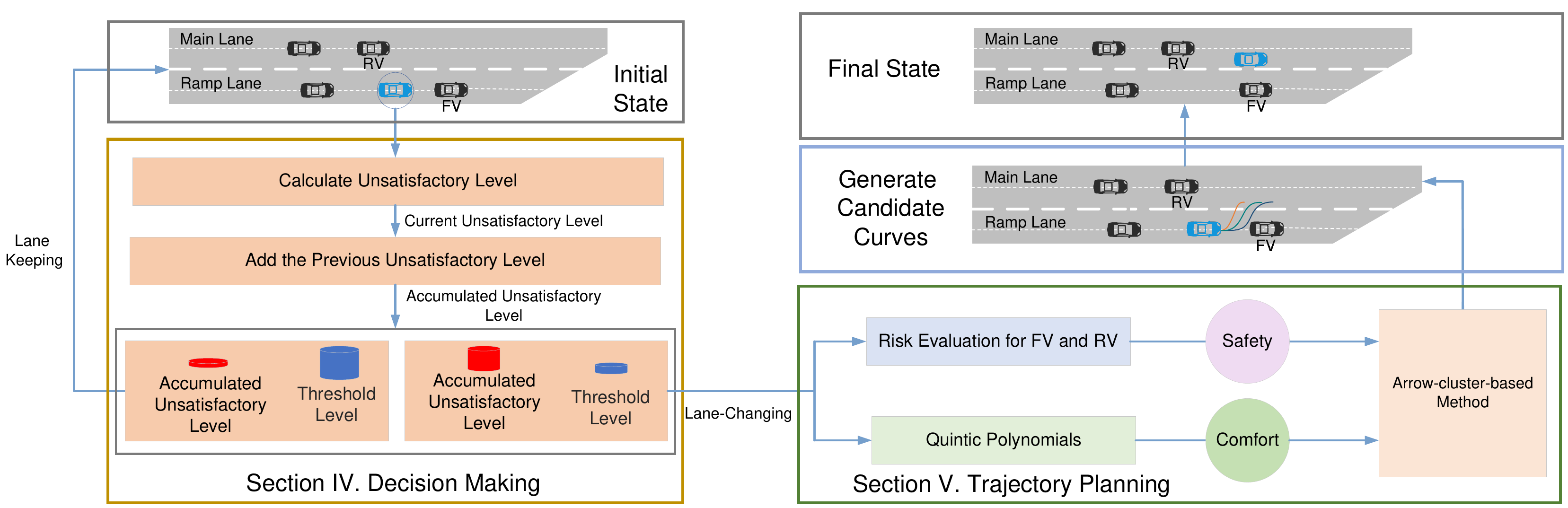}
    \caption{The proposed integrated unsatisfactory-based decision making and trajectory planning framework.}
    \label{fig2_interactive}
\end{figure*}
The rest of the paper is organized as follows: Section II introduces the related works. Section III introduces the overall architecture, and vehicle model. Section IV covers the unsatisfactory-based decision making. Section V presents the trajectory planning. Section VI reports the simulation results. Section VII draws the conclusions.

\section{Related Works}
\subsection{Driving Challenges in Ramp}
Driving on ramps is challenging due to the complex decisions, which involve merging, acceleration, deceleration, and potential conflicts with other vehicles. Ramp merging requires precise coordination between vehicles, and real-time decision-making to ensure the safety~\cite{zhu2022merging}. If a suitable opportunity for lane changing does not exist, the AV is expected to remain in the current lane, causing the congestion in ramp lane. During this phase, the AV must adjust its acceleration to maintain the safe space with front HDV. The Intelligent Driver Model (IDM) is widely used to model the longitudinal dynamics of the car-following process. However, the car-following phase faces the uncertainty of front HDV's motions, which disturbs the car-following process. Furthermore, the long-time remaining on the ramp lane will negatively affects the rear vehicles' driving. Therefore, the AV is required to makes lane-changing quickly.
\subsection{collision-avoidance with HDVs}
Various approaches have been proposed to address the collision-avoidance with dynamic obstacles~\cite{kamil2015review}.  Optimization-based methods, such as Model Predictive Control (MPC), have been increasingly applied to dynamic obstacle avoidance. While MPC can provide more globally optimal solutions compared to traditional methods, it is computationally expensive and may not be suitable for real-time applications. Another recent popular approach is artificial potential field (APF). APF uses attractive and repulsive forces to guide AVs toward target lanes while avoiding collisions~\cite{yao2020path}. However, it struggles to generalize across scenarios without prior environmental knowledge~\cite{triharminto2017local}.

For game-based method, Mont Carlo Tree Search (MCTS) has emerged as a promising alternative for dynamic obstacle avoidance, combining elements of game theory and probabilistic sampling~\cite{9082903}.  However, MCTS has an integrated process of decision-making and path-planning, decreasing the capability of inefficient decisions' correction during the lane-chaning. Therefore, MCTS is not robust enough for the safe and efficient lane-changing. 

The arrow-cluster-based method aims to balance the need for safe, efficient, and comfortable driving while interacting with HDVs in dynamic traffic environments. The method generates multiple potential trajectories based on safety, efficiency, and comfort. Compared to other planning methods, the arrow-cluster-based method provides greater flexibility and adaptability than rule-based methods~\cite{ulbrich2013probabilistic}. Besides, the arrow-cluster-based method does not require extensive training data like machine learning-based methods~\cite{tian2024balanced}. Moreover, the arrow-cluster-based method reduces computational complexity more effectively than optimization-based methods~\cite{frazzoli2002real}. 

\section{Problem Statement}
\subsection{The framework of proposed proposed planner}
Fig. 2 presents the proposed integrated planner for AV lane-changing on ramps with front vehicle (FV) and rear vehicl (RV), comprising Section IV and Section V: Decision Making and Trajectory Planning, respectively. In the Decision Making section, the system calculates an Unsatisfactory Level based on the current traffic state, accumulating it over time. When this Accumulated Unsatisfactory Level exceeds a predefined threshold, the system initiates a lane change. The Trajectory Planning section then generates candidate curves for the maneuver, evaluating them based on safety through risk assessment for the AV and surrounding vehicles, and comfort by using quintic polynomials. An Arrow-cluster-based Method is employed to select the optimal trajectory. The process begins with the AV in the ramp lane alongside other vehicles and concludes with the AV successfully merged into the main lane. This framework effectively integrates decision-making logic with trajectory planning, addressing both the timing of lane changes and their safe, comfortable execution.
\subsection{Vehicle model}
This paper assumes that the traffic flow on the ramp is normally high and the involved vehicles have relatively low speeds. To also simplify the computation during complex gaming,  the following kinematic model is used:
\begin{equation}
\dot{x} = v \cos(\varphi), \quad
\dot{y} = v \sin(\varphi), \quad
\dot{\varphi} = \frac{v \tan(\delta)}{L}
\label{eq2:vehicle_model}
\end{equation}
where \( x \) is the longitudinal position of the vehicle, \( y \) is the lateral position of the vehicle, \( \varphi \) is the orientation angle of the vehicle with respect to the x-axis, and \( L \) represents the wheelbase of the vehicle. The vehicle velocity \( v \) and the steering angle \( \delta \) are control inputs.
\section{Unsatisfactory-based Decision Making}
This section presents the decision process for lane keeping and lane-changing by using the unsatisfactory levels when driving in the ramp lane. The unsatisfactory level is modeled as a function of the difference between the desired speed and the actual speed of the vehicle. The unsatisfactory level enables the AV to make the decision to change lanes without hesitation, ensuring sufficient space in the ramp lane for other vehicles behind the AV.

\subsection{Unsatisfactory Level Calculation}
The primary measure of discomfort is computed as the percentage difference between the desired speed and the current speed of the AV. The discomfort at any time $t$ is given by:
\begin{equation}
    C_{\text{Per}}(t) = \frac{V_{\text{des}} - v(t)}{V_{\text{des}}} \cdot t
\end{equation}
where $V_{\text{des}}$ is the desired speed of the AV. $v(t)$ is the actual speed of the AV at time $t$. $t$ is the time point.
\subsection{Accumulated Unsatisfactory Level}
The accumulated satisfactory level is computed by integrating the discomfort over time. For the AV, this is expressed as:
\begin{equation}
    C_{\text{AV}}(t) = \sum_{i=0}^{t} \frac{V_{\text{des}} - V_{\text{AV}}(i)}{V_{\text{des}}} \cdot \Delta t
\end{equation}
where $\Delta t$ is the time step. Similarly, the discomfort for the main lane is calculated as:
\begin{equation}
    C_{\text{vir}}(t) = \sum_{i=0}^{t} \frac{V_{\text{des}} - V_{\text{vir}}(i)}{V_{\text{des}}} \cdot \Delta t
\end{equation}
where $V_{\text{vir}}(i)$ represents the speed of the vehicle in the main lane at time $i$.

\subsection{Double dimensional Sampling Curves}
This paper adopts sampling curves to generate a set of candidate paths. The goal of using the sampling methods is to select the best curve that meets the criteria of safety, comfort, and efficiency. The longitudinal distance of the end point $x_{e}$\, and the total time of the lane-changing process $t_{e}$, are usually within a relatively fixed range. This paper assumes that the longitudinal distance and time of lane-changing from the start point to the end point can be expressed by the following equation
\begin{equation}
\begin{aligned}
\left\{
\begin{array}{l}
x_{e,\min} = x_s + d_\text{min} \\
x_{e,\max} = x_s + d_\text{max}
\end{array}
\right.
\quad
\left\{
\begin{array}{l}
t_{e,\min} = t_s + T_\text{min} \\
t_{e,\max} = t_s + T_\text{max}
\end{array}
\right.
\end{aligned}
\end{equation}
where $x_{e,max}$\, and $x_{e,min}$\, are the maximum and minimum longitudinal coordinate respectively, $d_\text{max}$\, and $d_\text{min}$\, are the maximum and minimum longitudinal displacement respectively, $t_{e,max}$\, and $t_{e,min}$\, are the maximum and minimum time of lane changing respectively, $T_\text{max}$\, and $T_\text{min}$\, are the maximum and minimum time duration of lane changing respectively, $x_{s}$\, and $t_{s}$\, are the longitudinal coordinate and time of starting point respectively. Thus, the sampling range is formulated as follows
\begin{equation}
\left\{\begin{matrix} x_{e,range} =x_{e,min}:\,d_{s}:x_{e,max}\\ t_{e,range} =t_{e,min}:\,t_{s}:t_{e,max}\end{matrix}\right.
\end{equation}
where $d_{s}$\, and $t_{s}$\, are the displacement interval and time interval respectively. Assume that there are totally $M$\, number of $d_{s}$\, between $x_{e,min}$ to $x_{e,max}$\, and $N$\, number of $t_{s}$\, between $t_{e,min}$ to $t_{e,max}$\,, then the equations of the candidate paths from two dimension , $y-x$\, and $x-t$\,, is established as follows
$$ S_{y-x}(m)=a_{0}(m)+a_{1}(m) x+ a_{2}(m)x^{2}+a_{3}(m)x^{3}+a_{4}(m)x^{4} $$
$$ + a_{5}(m)x^{5}$$
$$ m=1,2,..., M$$
$$  s.t. :if \,\, x=0 : y(m)=y_{s},\dot{y}(m)=tan(\phi _{s}), $$
$$ \ddot{y}(m)=(1+\dot{y}_{m}^2)^\frac{3}{2}tan\delta _{s}L^{-1}  $$
$$  if \,\,x=x_{e}(m) : y(m)=y_{s}+w,\dot{y}(m)=0,\ddot{y}(m)=0 $$
$$ \, $$
$$ S_{x-t}(n)=b_{0}(n)+b_{1}(n) t+ b_{2}(n)t^{2}+b_{3}(n)t^{3}+b_{4}(n)t^{4} $$
$$ + b_{5}(n)t^{5}$$
$$ n=1,2,..., N$$
$$  s.t. :if \,\, t=0 : x(n)=0,\dot{x}(n)=V_{s}cos(\phi _{s}), $$
$$ \ddot{x}(n)=a_{s}cos{\phi_{s}}+v_{s}^2tan\delta _{s}sin\phi _{s}L^{-1}  $$
\begin{equation}
 if \,\,t=t_{e}(n) : x(n)=x_{e}(m),\dot{x}(n)=v_\text{comfort},\ddot{x}(n)=0 
\end{equation}
where $a_{0}$\,,$a_{1}$\,,$a_{2}$\,,$a_{3}$\,,$a_{4}$\, and $a_{5}$\, are the coefficients of $S_{y-x}$\,, $v_{s}$\,,$a_{s}$\,, $\delta_{s}$\, and $\delta_{s}$\, are the starting velocity, starting acceleration, starting yaw angle and starting steering angle respectively, $b_{0}$\,,$b_{1}$\,,$b_{2}$\,,$b_{3}$\,,$b_{4}$\, and $b_{5}$\, are the coefficients of $S_{x-t}$\,, $L$\, and $w$\, are the wheel base and width respectively. $v_{comfort}$\, is the comfortable velocity of driving. Using (9), a series of candidate paths can be generated to provide sufficient samples for optimizing the lane-changing curve.
\subsection{Loss Function for Candidates paths}
To select a path that best fit the requirement of safety, efficiency and comfort, a loss function is proposed as follows
\begin{equation}
U_{total}=w_{y-x}U_{y-x}+w_{x-t}U_{x-t}+w_\text{\text{risk}}U_\text{\text{risk}}  
\end{equation}
where $U_{y-x}$\,, $U_{x-t}$\, and $U_\text{\text{risk}}$\, are the loss of curves $y-x$\,,curve $x-t$\, and hybrid risk filed, $w_{y-x}$\,, $w_{x-t}$\, and  $w_\text{\text{risk}}$\, are the weight factors of $U_{y-x}$\,, $U_{x-t}$\, and $U_\text{\text{risk}}$\, respectively. $U_{y-x}$\, is formulated by the following equation
$$ U_{y-x}=w_{y-x,1}\int_{0}^{x_{e}(m)}\dot{S}_{y-x}^2dx+ w_{y-x,2}\int_{0}^{x_{e}(m)}\ddot{S}_{y-x}^2dx $$
\begin{equation}
+w_{y-x,3}\int_{0}^{x_{e}(m)}\dddot{S}_{y-x}^2dx    
\end{equation}
where $\int_{0}^{x_{e}(m)}\dot{S}_{y-x}^2 \, dx$, $\int_{0}^{x_{e}(m)}\ddot{S}_{y-x}^2 \, dx$, and $\int_{0}^{x_{e}(m)}\dddot{S}_{y-x}^2 \, dx$ are the length of the curves, the curvature of the curve length, and the curvature variation of the curve length, respectively,  $w_{y-x,1}$\,, $w_{y-x,2}$\, and $w_{y-x,3}$\, are the weight factors of $\int_{0}^{x_{e}(m)}\dot{S}_{y-x}^2dx$\,, $\int_{0}^{x_{e}(m)}\ddot{S}_{y-x}^2dx$\, and $\int_{0}^{x_{e}(m)}\dddot{S}_{y-x}^2dx$\,. This component can reflect the smoothness of curves from the dimension of $y-x$\,.  $U_{x-t}$\, is formulated by the following equation  
$$ U_{x-t}=w_{x-t,1}\int_{0}^{t_{e}(n)}(\dot{S}_{x-t}-V_\text{des})^2dt+ w_{x-t,2}\int_{0}^{t_{e}(n)}\ddot{S}_{x-t}^2dt $$
\begin{equation}
 +w_{x-t,3}\int_{0}^{t_{e}(n)}\dddot{S}_{x-t}^2dt   
\end{equation}
where $\int_{0}^{x_{e}(n)}(\dot{S}_{x-t}-V_\text{des})^2dt$\,, $\int_{0}^{x_{e}(n)}\ddot{S}_{x-t}^2dt$\, and $\int_{0}^{x_{e}(n)}\dddot{S}_{x-t}^2dt$\, are the velocity difference, the curvature of velocity, and the curvature variation of velocity, respectively,  $w_{x-t,1}$\,, $w_{x-t,2}$\, and $w_{x-t,3}$\, are the weight factors of $\int_{0}^{x_{e}(n)}\dot{S}_{x-t}^2dt$\,, $\int_{0}^{x_{e}(n)}\ddot{S}_{x-t}^2dt$\, and $\int_{0}^{x_{e}(n)}\dddot{S}_{x-t}^2dt$\,. This component can assess the smoothness of curves from the dimension of $x-t$\,.  $U_{\text{risk}}$\, is formulated by the following equation 
\begin{equation}
\begin{split}
 U_{\text{risk}} &= \sum_{i=1}^{2} \frac{1}{1 - e^{-d_{\text{risk}_i}}} \\[10pt]
\text{where } d_{\text{risk}_1} &= p_1 \Delta y_1^2 + p_2 \Delta x_1^2, \\[5pt]
\Delta y_1 &= y_\text{AV} - y_\text{RV}, \quad \Delta x_1 = x_\text{AV} - x_\text{RV}, \\[10pt]
d_{\text{risk}_2} &= \min_{j} \left[ p_1 \Delta y_{2,j}^2 + p_2 \Delta x_{2,j}^2 \right], \\[5pt]
\Delta y_{2,j} &= y_{AV,j} - y_\text{FV}, \quad \Delta x_{2,j} = x_{AV,j} - x_{FV,j}
\end{split}
\end{equation}

The risk loss function $S_{ob}$ calculates the total risk as the sum of two inverse exponential terms, each representing the risk associated with one of two obstacles. where $(x_\text{AV}, y_\text{AV})$ denotes the position of the host vehicle, $(x_\text{RV}, y_\text{RV})$ is the position of RV, and $(x_{\text{FV},j}, y_\text{FV})$ represents the position of FV at time step j. The individual risk terms $d_{\text{\text{risk}}_1}$ and $d_{\text{risk}_2}$ are computed based on the weighted squared Euclidean distances between the AV, RV and FV, with $p_1$ and $p_2$ serving as weighting parameters, respectively. 
\begin{figure}[h]
    \centering
    \includegraphics[width=0.7\linewidth]{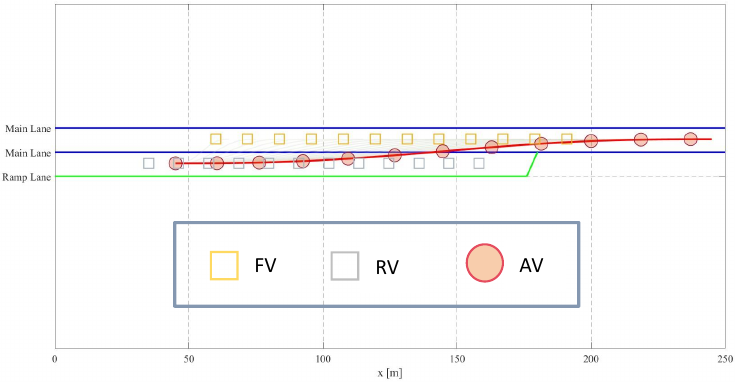}
    \vspace{-2mm}
    \caption{The driving performance in the simulated ramp.}
    \label{fig1_framework}
\end{figure}
\begin{figure}[h]
    \centering
    \includegraphics[width=1\linewidth]{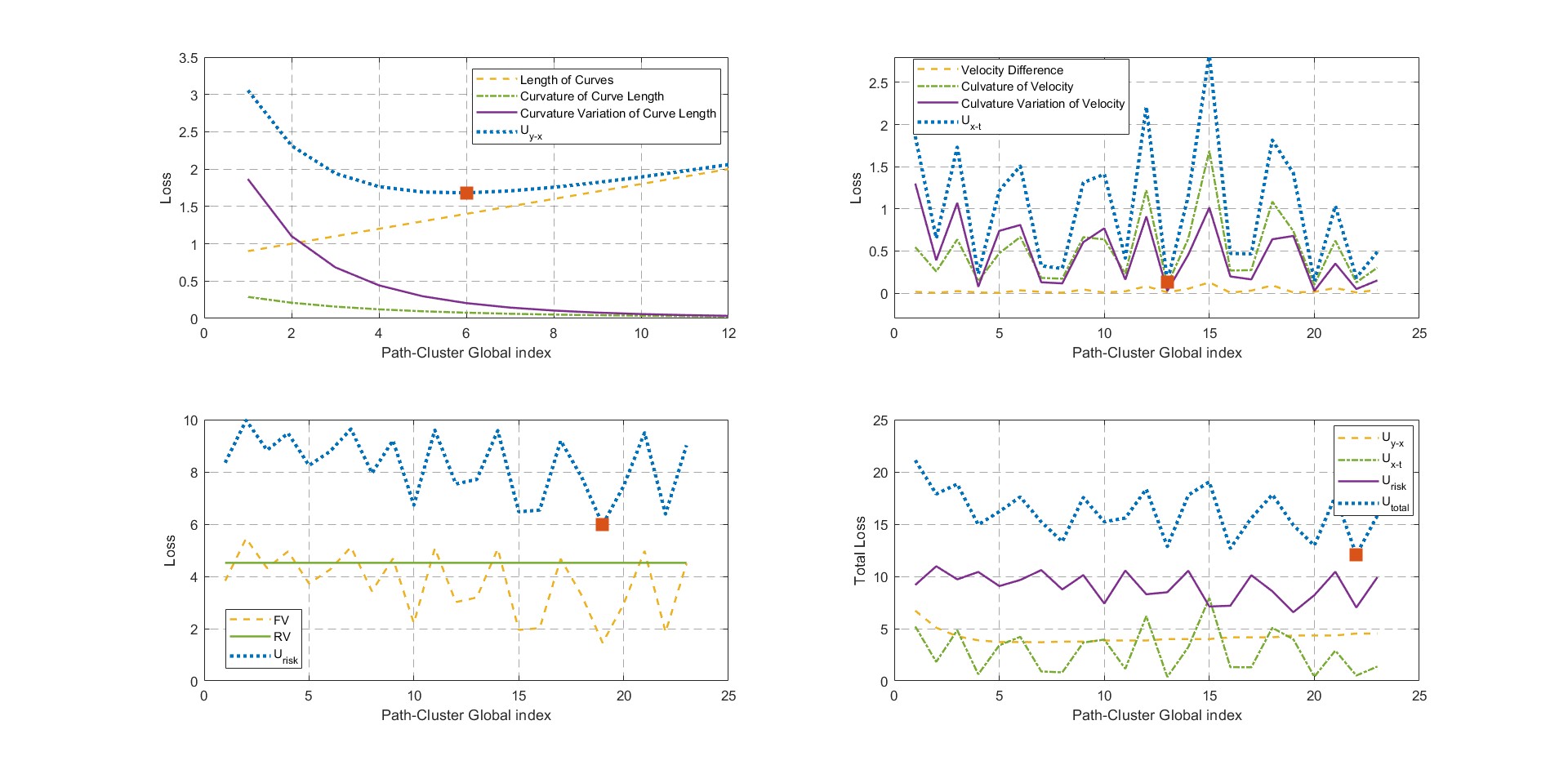}
    \vspace{-2mm}
    \caption{The loss curves of $U_{x-y}$,$U_{x-t}$, $U_\text{risk}$, and $U_\text{total}$.}
    \label{fig1_framework}
\end{figure}
\section{Simulation Results}

\begin{figure*}[h]
    \centering
    \includegraphics[width=0.6\linewidth]{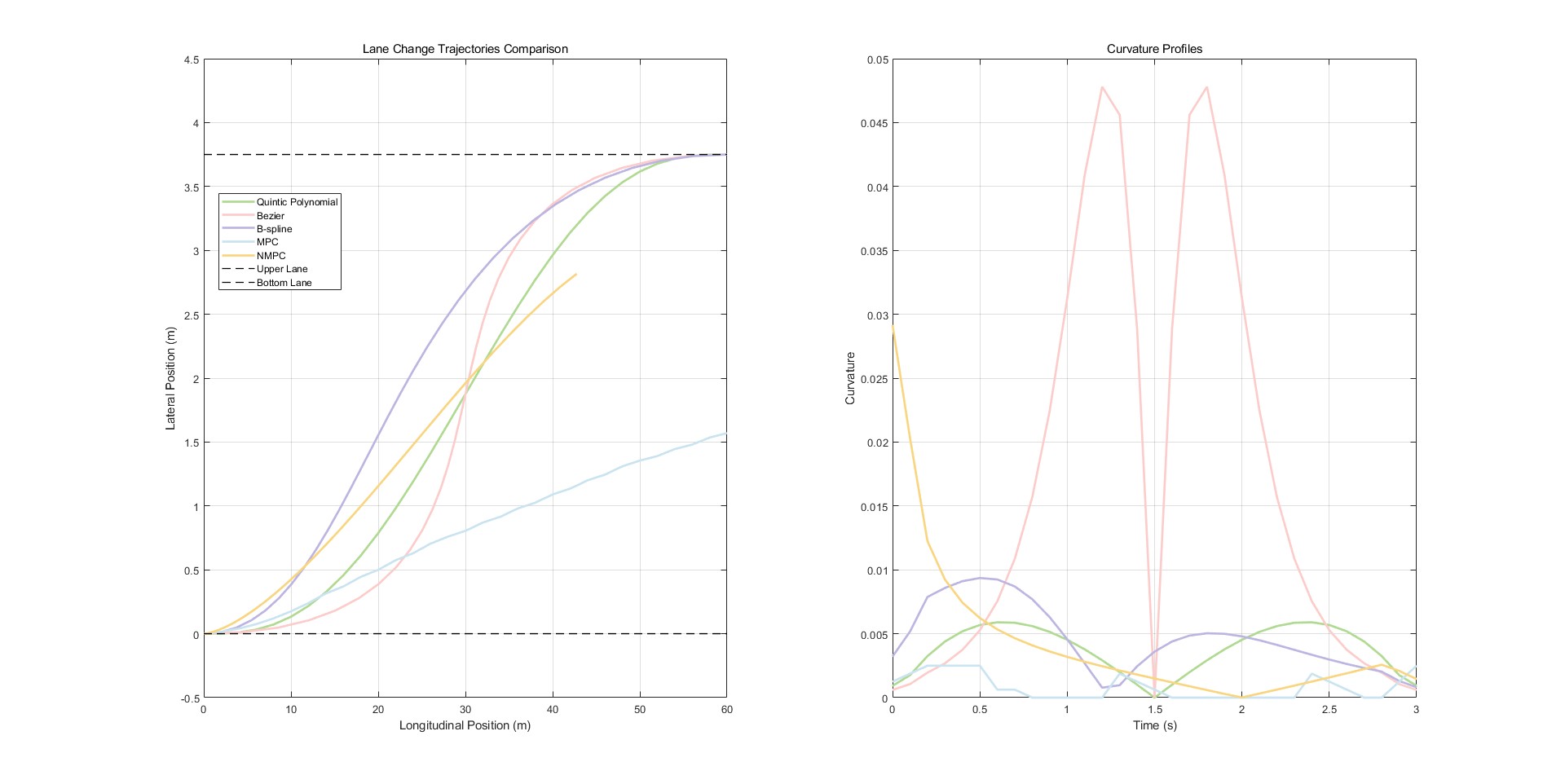}
    \vspace{-2mm}
    \caption{Comparison of lane-changing trajectories and corresponding curvature profiles for different methods}
    \label{fig1_framework}
\end{figure*}

\begin{figure}[t]
    \centering
    \includegraphics[width=0.5\linewidth]{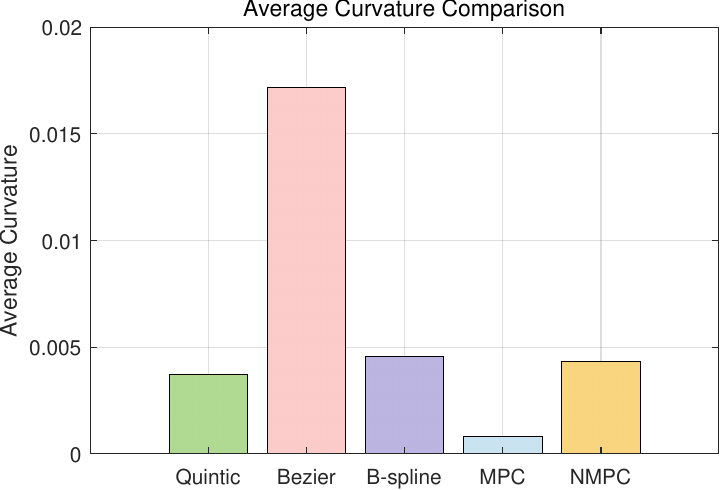}
    \vspace{-2mm}
    \caption{Comparison of average curvature values for different trajectory generation methods.}
    \label{fig1_framework}
    \end{figure}
This section presents our experimental results and an analysis of the proposed on-ramp merging framework in Matlab 2022b. The simulations aim to assess the proposed framework by examining exploring whether the AV merging from the ramp onto the main road with efficiency, safety and comfort. There are totally two lanes in the simulation, including a ramp lane and a main lane; the length and width of the main lane are 250 and 3.5 m;the length and width of the ramp lane are 180 and 3.5 m  the length and width of vehicles are 4.5 and 2 m, respectively; the time resolution is 0.1 s; the vehicles flow on the ramp lane is 200 vehicles/h; the vehicles flow on the straight lane is 100 vehicles/h; the initial velocity of AV is 22 m/s, the $v_\text{des}$ is 26,4 m/s; the max and min velocity of vehicles are 33 m/s and 0 m/s. 

Fig. 2 shows the AV’s lane-change trajectory from the ramp to the main lane over a 0–250 m horizontal span. The two main lanes (blue) and the ramp lane (green) are depicted, with vehicles represented as follows: the AV (red circles), the FV (yellow squares ahead in the target lane), and the RV (gray squares following in the ramp lane). The AV’s trajectory is highlighted by a red curve, and the candidate S-shaped curves are also shown; the selected curve is smooth and collision-free.

Fig. 3 analyzes key path planning metrics in four subplots. The top-left subplot presents the lateral path quality ($U_{y-x}$), showing a convex curve with a clear minimum. The top-right subplot displays the longitudinal velocity profile ($U_{x-t}$) with notable fluctuations. The bottom-left subplot focuses on risk assessment ($U_{\text{risk}}$), revealing significant variability in collision risk. Finally, the bottom-right subplot illustrates the total loss ($U_{\text{total}}$) along with its components, highlighting multiple local minima that reflect trade-offs among lateral movement, velocity control, and risk.

Fig. 4 compares lane-change trajectories and curvature profiles from different methods, including quintic polynomial (ours), Bezier~\cite{10208112}, B-spline~\cite{9415170}, MPC~\cite{9547835}, and NMPC~\cite{ismael2024nonlinear}. Our quintic polynomial method demonstrates a smooth, continuous trajectory that ensures safety and comfort, while the Bezier and B-spline methods show steeper lateral changes. Although MPC and NMPC also yield reasonable smoothness, our method provides a more gradual and stable transition.
    \begin{figure}[t]
    \centering
    \includegraphics[width=0.6\linewidth]{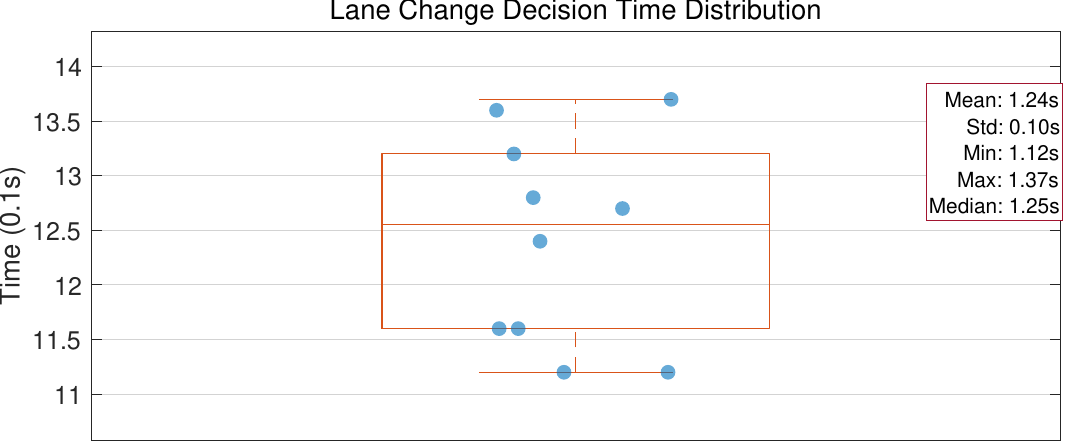}
    \vspace{-2mm}
\caption{Lane-changing decision time distribution for varying initial relative distances with the FV within a range of -15 to 15 meters, repeated over 10 trials.}
    \label{fig1_framework}
    \end{figure}

    \begin{figure}[t]
    \centering
    \includegraphics[width=0.6\linewidth]{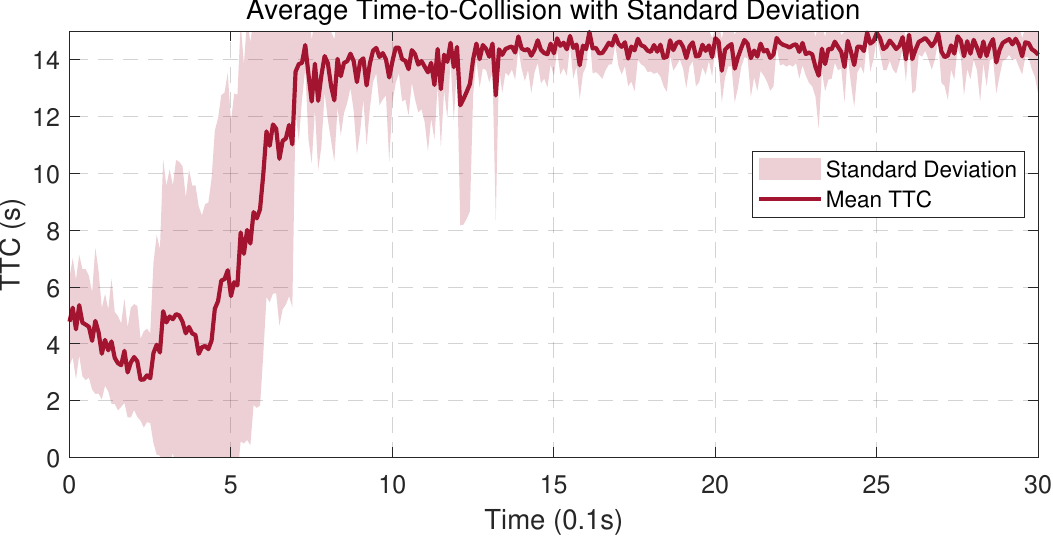}
    \vspace{-2mm}
\caption{TTC distribution for varying initial relative distances within a range of -15 to 15 meters, repeated over 10 trials.}
    \label{fig1_framework}
    \end{figure}

       \begin{figure}[t]
    \centering
    \includegraphics[width=0.5\linewidth]{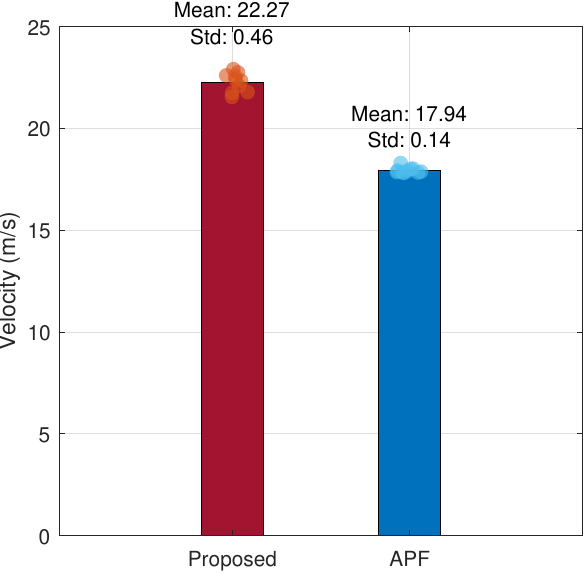}
    \vspace{-2mm}
\caption{Comparison of average velocity distribution between the proposed method and APF for varying initial relative distances with the FV within a range of -15 to 15 meters, repeated over 10 trials.}
    \label{fig1_framework}
    \end{figure}
The right panel presents the curvature profiles for these trajectories. The quintic polynomial method achieves a moderate curvature variation, reflecting smooth transitions in steering. In contrast, the Bezier method shows significant peaks in curvature, indicating sharp turns that may compromise passenger comfort. Meanwhile, the curvature of B-spline, MPC, and NMPC methods exhibits noticeable variability, particularly in the initial and middle stages of the lane change. Fig. 6 presents a comparison of the average curvature values for different trajectory generation methods. This indicates that the Bezier method generates trajectories with sharper turns, which may compromise passenger comfort and vehicle stability during lane changes. In contrast, the quintic polynomial method achieves a second-lowest average curvature. This highlights the quintic polynomial’s ability to produce smooth and gradual trajectory transitions, ensuring a balance between safety and comfort..

Fig. 6 illustrates the distribution of lane change decision times, with a mean of $1.24~\text{s}$, median of $1.25~\text{s}$, and a standard deviation of $0.10~\text{s}$, indicating relatively consistent and fast decision-making. The narrow range, spanning from $1.12~\text{s}$ to $1.37~\text{s}$, suggests that drivers under unsatisfactory orientation, tend to make lane-change decisions fast.

Time-to-Collision (TTC) is a key safety metric that estimates the time before a potential collision~\cite{zhao2023effects}. Higher TTC values indicate safer conditions, while low or negative values signal potential risks. The $3-\text{s}$ TTC threshold is widely acknowledged in peer-reviewed literature as a standard warning criterion for collision avoidance systems~\cite{das2019defining}. Fig. 8 illustrates the average TTC between the AV and the FV, along with the standard deviation over time. At approximately $0.5~\text{s}$, the TTC briefly drops below zero; however, the FV is on the adjacent lane during this phase, thereby ensuring no actual collision occurs. By around $1.24~\text{s}$, the average lane-changing decision time, the AV adjusts above a safe $3-\text{s}$ TTC before initiating the lane change, demonstrating the safety and intelligence of the proposed method. The TTC stabilizes at approximately $1.4~\text{s}$ after $0.75~\text{s}$, with significantly reduced fluctuations in standard deviation, reflecting consistent and safe vehicle spacing post-lane-change.

Fig. 9 compares the average velocity of the proposed method with the APF approach. The proposed method achieves a significantly higher mean velocity of $22.27~\text{m/s}$, substantially outperforming the APF method, which achieves only $17.94~\text{m/s}$. The standard deviation for the proposed method is $0.46~\text{m/s}$, slightly higher than the APF method's $0.14~\text{m/s}$, reflecting a minor trade-off for the significant improvement in velocity.
\section{conclusion}
This paper presents a comprehensive decision making framework for autonomous vehicles in ramp lane change scenarios, integrating an unsatisfactory level indicator with an arrow cluster based planning approach. Our framework demonstrates significant advantages through three key innovations: a metric for unsatisfactory level that promptly signals when a lane change maneuver is necessary; an arrow cluster based planning method that robustly evaluates future vehicle state losses to select the most efficient, safe, and comfortable trajectory; and an integrated optimization strategy that harmonizes reactive and predictive decision making. Experimental results in various ramp scenarios show that our approach triggers lane change intentions quickly while producing smooth, collision free trajectories (with a time to collision, TTC, consistently exceeding the safe 3 s threshold) and achieves higher average speeds than benchmark collision avoidance methods. Future work will focus on enhancing computational efficiency and extending the framework to accommodate more complex ramp geometries and mixed traffic conditions.

\bibliographystyle{IEEEtran}
\bibliography{IEEEabrv,zq_lib}

\end{document}